# TSAX is Trending

Muhammad Marwan Muhammad Fuad

Coventry University
Coventry CV1 5FB, UK
ad0263@coventry.ac.uk

**Abstract:** Time series mining is an important branch of data mining, as time series data is ubiquitous and has many applications in several domains. The main task in time series mining is classification. Time series representation methods play an important role in time series classification and other time series mining tasks. One of the most popular representation methods of time series data is the Symbolic Aggregate approXimation (SAX). The secret behind its popularity is its simplicity and efficiency. SAX has however one major drawback, which is its inability to represent trend information. Several methods have been proposed to enable SAX to capture trend information, but this comes at the expense of complex processing, preprocessing, or post-processing procedures. In this paper we present a new modification of SAX that we call Trending SAX (TSAX), which only adds minimal complexity to SAX, but substantially improves its performance in time series classification. This is validated experimentally on 50 datasets. The results show the superior performance of our method, as it gives a smaller classification error on 39 datasets compared with SAX.

## 1 Introduction

Time series mining has witnessed substantial interest in the last two decades because of the popularity of time series data, as much of the data in the world is in the form of time series [15].

Time series mining deals with several tasks such as classification, clustering, segmentation, query-by-content, anomaly detection, prediction, and others.

Time series classification (TSC) is the most important time series mining task. TSC has a wide variety of applications in medicine, industry, meteorology, finance, and many other domains. This variety of applications is the reason why TSC has gained increasing attention over the last two decades [2] [4] [6] [8] [22].

The most common TSC method is the *k-nearest-neighbor* (kNN), which applies the similarity measures to the object to be classified to determine its best classification based on the existing data that has already been classified. The performance of a classification algorithm is measured by the percentage of objects identified as the correct class [15]. kNN can be applied to raw time series or to lower-dimension representations of the time series. These representation methods are widely used in time series mining as time series may contain noise or outliers. Besides, performing TSC on lower-dimension representations of time series is more efficient than applying it directly to raw data. The most widely used kNN method is 1NN.

There have been a multitude of time series representation methods in the literature, to name a few; the Piecewise Aggregate Approximation (PAA) [9] [23], Adaptive Piecewise Constant Approximation (APCA) [10], and the Clipping Technique [21].

The Symbolic Aggregate Approximation (SAX) [11] [12] is one of the most popular time series representation methods. The secret behind this popularity is its efficiency and simplicity. SAX is in fact based on PAA, but it applies further steps so that the time series is converted into a sequence of symbols.

Researchers have, however, pointed to a certain drawback in SAX, which is its inability to capture trend information. This is an important feature in many TSC applications.

In this work we present a new modification of SAX that captures the trend information with very small additional storage and computational costs compared with the original SAX. The extensive experiments we conducted on a wide variety of time series datasets in a TSC task show a substantial boost in performance.

The rest of the paper is organized as follows: Section 2 discusses background material on time series data, mainly SAX. Section 3 introduces our new method TSAX. In Section 4 we conduct extensive experiments that compare TSAX against SAX. We conclude with Section 5.

## 2 Background

A time series $T = (t_1, t_2, ..., t_n)$ is an ordered collection of $n$ measurements at timestamps $t_n$. Time series data appear in a wide variety of applications.

Given a time series dataset $D$ of $s$ time series, each of $n$ dimensions. Each time series $T_i$, $i \in \{1,2,...,s\}$ is associated with a class label $L(T_i)$; $L(T_i) \in \{1,2,...,c\}$. Given a set of unlabeled time series $U$, the purpose of the TSC task is to map each time series in $U$ to one of the classes in $\{1,2,...,c\}$. TSC algorithms involve some processing or filtering of the time series values prior or during constructing the classifier. The particularity of TSC, which makes it different from traditional classification tasks, is the natural temporal ordering of the attributes [1].

$1NN$ with the Euclidean distance or the Dynamic Time Warping (DTW), applied to raw time series, has been widely used in TSC, but the Euclidean distance is weak in terms of accuracy [20], and DWT, although much more accurate, is computationally expensive.

Instead of applying different tasks directly to raw time series data, time series representation methods project the data on lower-dimension spaces and perform the different time series mining tasks in those low-dimension spaces.

A large number of time series representation methods have been proposed in the literature. The Piecewise Aggregate Approximation (PAA) [9] [23] is one of the first and most simple, yet effective, time series representation methods in the literature.

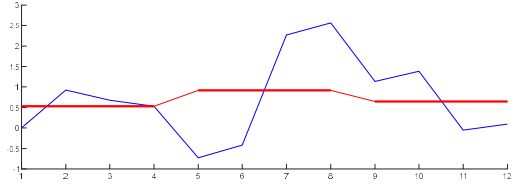
**Fig. 1.** PAA representation

PAA divides a time series $T$ of $n$-dimensions into $m$ equal-sized segments and maps each segment to a point of a lower $m$-dimensional space, where each point in this space is the mean of values of the data points falling within this segment as shown in Fig. 1.

PAA is the predecessor of another very popular time series representation method, which is the Symbolic Aggregate Approximation – SAX [11] [12]. SAX performs the discretization by dividing a time series $T$ into $m$ equal-sized segments (words). For each segment, the mean value of the points within that segment is computed. Aggregating these $m$ coefficients forms the PAA representation of $T$. Each coefficient is then mapped to a symbol according to a set of breakpoints that divide the distribution space into $\alpha$ equiprobable regions, where $\alpha$ is the *alphabet size* specified by the user. The locations of the breakpoints are determined using a statistical lookup table for each value of $\alpha$. These lookup tables are based on the assumption that normalized time series subsequences have a highly Gaussian distribution [12].

It is worth mentioning that some researchers applied optimization, using genetic algorithms and differential evolution, to obtain the locations of the breakpoints [16] [18]. This gave better results than the original lookup tables based on the Gaussian assumption.

To summarize, SAX is applied to normalized time series in three steps as follows:

1- The dimensionality of the time series is reduced using PAA.

2- The resulting PAA representation is discretized by determining the number and locations of the breakpoints. The number of the breakpoints $nrBreakPoints$ is: $nrBreakPoints = \alpha - 1$. As for their locations, they are determined, as mentioned above, by using Gaussian lookup tables. The interval between two successive breakpoints is assigned to a symbol of the alphabet, and each segment of PAA that lies within that interval is discretized by that symbol.

3- The last step of SAX is using the following distance:

$$MINDIST(\hat{S}, \hat{T}) = \sqrt{\frac{n}{m}} \sqrt{\sum_{i=1}^{m} \left(dist(\hat{s}_i, \hat{t}_i)\right)^2} \qquad (1)$$

Where $n$ is the length of the original time series, $m$ is the number of segments, $\hat{S}$ and

Table 1. The lookup table of *MINDIST* for alphabet size = 4.

|   | a    | b    | c    | d    |
|---|------|------|------|------|
| a | 0    | 0    | 0.67 | 1.34 |
| b | 0    | 0    | 0    | 0.67 |
| c | 0.67 | 0    | 0    | 0    |
| d | 1.34 | 0.67 | 0    | 0    |

$\hat{T}$ are the symbolic representations of the two time series $S$ and $T$, respectively, and where the function $dist(\ )$ is implemented by using the appropriate lookup table. For instance, the lookup table for an alphabet size of 4 is the one shown in Table 1. Fig. 2 shows an example of SAX for $\alpha = 4$

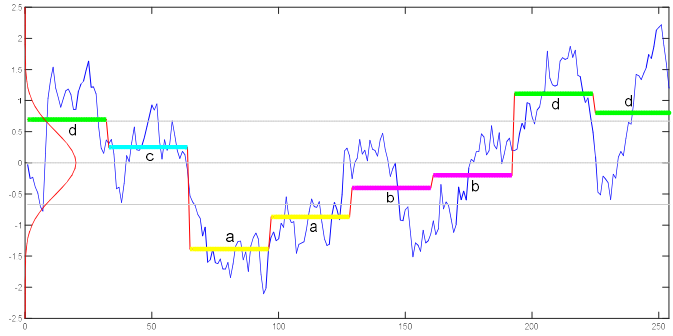

**Fig. 2.** Example of SAX for $\alpha = 4, m = 8$. In the first step the time series, whose length is 256, is discretized using PAA, and then each segment is mapped to the corresponding symbol. This results in the final SAX representation for this time series, which is $dcaabbdd$

## 3 Trending SAX (TSAX)

**3.1 Motivation and Related Work**

The efficiency and simplicity of SAX made of it one of the most popular time series representation methods. But SAX has a main drawback, which is its lack of a mechanism to capture the trends of segments. For example, the two segments: $S_1 = [-6, -1, +7, +8]$ and $S_2 = [+9, +3, +1, -5]$ have the same PAA coefficient which is $+2$, so their SAX representation is the same, although, as we can clearly see, their trends are completely different.

Several researchers have reported this flaw in SAX and attempted to propose different solutions to handle it. In [14] the authors present 1d-SAX which incorporates trend information by applying linear regression to each segment of the time series and then discretizing it using a symbol that incorporates both trend and mean of the segment at the same time. The method applies a complex representation and training scheme. In [24] another extension of SAX that takes into account both trend and value is proposed. This method first transforms each real value into a binary bit. This representation is used in a first pruning step. In the second step the trend is represented as a binary bit. This method requires two passes through the data which makes it inefficient for large datasets. [7] proposes to consider both value and trend information by combining PAA with an extended version of the clipping technique [21]. However, obtaining good results requires prior knowledge of the data. In [17] and [19] two methods to add trend information to SAX are proposed. The two methods are quite simple, but the improvement is rather small.

In addition to all these drawbacks we mentioned above of each of these methods, which attempt to enable SAX to capture trend information, the main disadvantage of these methods (except [17] and [19]) is that they apply a complex representation and training method, whereas the main merit of SAX is its simplicity with no need for any training stage.

### 3.2 TSAX Implementation

Our proposed method, which we call *Trending SAX* (TSAX), has the same simplicity as the original SAX. It has an additional, but quite small, storage and computational requirement. It has extra features that enable it to capture segment trend information and it uses this trend information when comparing two time series for TSC, which enables it to obtain substantially better results. Compared with SAX, TSAX has two additional parameters. This might seem as a downside to using our new method, as parameters usually require training. But our method can be used without any training of the parameters, as we will show later in the experimental section, so as such, these parameters are very much like constants. In fact, we wanted to test our method under strictly unbiased conditions, so we gave these parameters rather arbitrary values, yet the performance was superior to that of SAX, as we will see in the experiments.

Before we introduce our method, we have to say that although in this paper we apply it to TSC only, the method can be applied to other time series mining tasks.

In the following we present the different steps of TSAX: Let $T$ be a time series in a $n$-dimensional space to be transformed by TSAX into a $2m$-dimensional space. After normalization, $T$ is segmented into $m$ equal- sized segments (words) $w_i$, $i = 1, \dots, m$, where each word $w_i$ is the mean of the data points lying within that segment (the same steps of PAA described above). In the next step each segment is represented by the corresponding alphabet symbol exactly as in SAX, where $\alpha$, the alphabet size, is determined by the user.

In the following step, for each word $w_i$, the trend $tr_i$ of the corresponding segment of time series raw data is calculated by first computing the linear regression of the time series raw data of $w_i$, then $tr_i$ is represented as ↗ if it is an upward trending line or by ↘ if it is a downward trending line, so after executing all the above steps, each time series $T$ will be represented in TSAX as:

$$T \rightarrow [w_1 w_2 w_3 \dots w_m \; tr_1 tr_2 tr_3 \dots tr_m] \qquad (2)$$

Where $tr_i$ takes one of the values ↗ or ↘ depending on whether the corresponding linear regression line of the raw time series corresponding to $w_i$ is an upward or downward trend, respectively. We call the first $m$ components of the above representation $[w_1 w_2 w_3 \ldots w_m]$ the *symbolic part* of the TSAX representation, whereas $[tr_1 tr_2 tr_3 \ldots tr_m]$ is called the *trend part* of the TSAX representation.

Notice that the values of the end points of the linear regression line of each segment are real numbers, so the probability that they are equal is zero, so the trend is either ↗ or ↘. However, because of the way real number are represented on a computer, the two end points could theoretically have the same value, so, for completion, we can choose to represent this (very) special case, where the two values are equal, as either ↗ or ↘. But this does not have any practical importance.

Also notice, which is important as we will show later, that trend has two values only, so it can be represented by a Boolean variable.

Let $T, T'$ be two time series whose TSAX representations are:
$[w_1 w_2 w_3 \ldots w_m\ tr_1 tr_2 tr_3 \ldots tr_m]$ and $[w_1' w_2' w_3' \ldots w_m'\ tr_1' tr_2' tr_3' \ldots tr_m']$, respectively. The symbolic parts of the two representations are compared the same way as in SAX, as we will show later. The trend parts are compared as follows: each of $tr_i$ is compared to its counterpart $tr_i'$, $i = 1, \ldots, m$ according to the following truth table:

| $tr_i$ | $tr_i'$ | *trendMatch* |
|---|---|---|
| ↗ | ↗ | == |
| ↘ | ↗ | ≠≠ |
| ↗ | ↘ | ≠≠ |
| ↘ | ↘ | == |

Where *trendMatch* is also represented by a Boolean variable. The symbol (==) means that the two trends $tr_i$ and $tr_i'$ are the same, whereas the symbol (≠≠) means that the two trends $tr_i$ and $tr_i'$ are opposite.

Finally, the TSAX distance between the time series $T, T'$, whose TSAX representations are denoted $\hat{T}, \hat{T}'$, respectively, is:

$$TSAX\_DIST(\hat{T}, \hat{T}') = \sqrt{\frac{n}{m}} \sqrt{\sum_{i=1}^{m} \left(dist(\hat{t}_i, \hat{t}_i')\right)^2} + rew \times k_1 + pen \times k_2$$

(3)

Where $k_1$ is the number of times the variable *trendMatch* takes the value (==), i.e. the number of times the two time series have segments with matching trends (to their counterparts in the other time series), and $k_2$ is the number of times the variable *trendMatch* takes the value (≠≠), i.e. the number of times the two time series have segments with opposite trends. *rew* (stands for "reward") and *pen* (stands for "penalty") are two parameters of our method. We will discuss how they are selected in the experimental section of this paper.

Notice that equation (3) is a distance, so when two time series have more segments with matching trends it means the two time series should be viewed as "closer" to each other, i.e. *rew* should take a negative value, whereas *pen* takes a positive value.

### 3.3 Illustrating Example

In order to better explain our method, we give an example to show how it is applied. Let $T, T'$ be the following two time series of length 16, i.e. $n = 16$ in equation (3):

$T = [-7.1 \ -1.1 \ -1.3 \ -1.5 \ -1.4 \ -1.3 \ -1.0 \ 4.5 \ 9.2 \ 1.0 \ 1.2 \ 9.6 \ 6.1 \ 1.4 \ -6.4 \ -2.6]$
$T' = [-9.9 \ -1.4 \ -1.5 \ -1.6 \ -1.6 \ -1.3 \ -1.0 \ -3.5 \ 7.1 \ 1.2 \ 1.1 \ 1.0 \ 7.9 \ 4.6 \ 4.8 \ 5.6]$

After normalization (values are rounded up because of space limitations):

$T \rightarrow [-1.6 \ -0.4 \ -0.4 \ -0.4 \ -0.4 \ -0.4 \ 0.3 \ 0.8 \ 1.8 \ 0.1 \ 0.1 \ 1.9 \ 1.2 \ 0.2 \ -1.5 \ -0.7]$
$T' \rightarrow [-2.4 \ -0.5 \ -0.5 \ -0.5 \ -0.5 \ 0.4 \ -0.4 \ -1.0 \ 1.4 \ 0.1 \ 0.1 \ 0.1 \ 1.6 \ 0.9 \ 0.9 \ 1.1]$

If we choose $\frac{n}{m} = 4$, which is the value used in the original SAX, and the one used in the experimental section of this paper, then $m = 4$. The PAA representations, where each component is the mean of four successive values of $T, T'$, are the following:

$$T\_PAA \rightarrow [-0.700 \ \ 0.075 \ \ 0.975 \ -0.200]$$
$$T'\_PAA \rightarrow [-0.975 \ -0.375 \ \ 0.425 \ \ 1.125]$$

For alphabet size $\alpha = 4$ (the one used in the experiments), the symbolic parts of the TSAX representations are the following:

$$T\_TSAX\_SymbolicPart = [acdb]$$
$$T'\_TSAX\_SymbolicPart = [abcd]$$

As for the trend part, we need to compute the linear regression of each segment of the raw date. For instance, the linear regression of segment $[-1.6 \ -0.4 \ -0.4 \ -0.4]$ is $[-1.2206 \ -0.8725 \ -0.5245 \ -0.1764]$, so the trend for this segment of $T$ is ↗. Continuing in the same manner for the other segments we get:

$$T\_TSAX\_TrendPart = [↗ ↗ ↗ ↘]$$
$$T'\_TSAX\_TrendPart = [↗ ↘ ↘ ↘]$$

When comparing the trend match between $T$ and $T'$ we get: $[== \neq\neq \ \neq\neq \ ==]$, so $k_1 = 2$ and $k_2 = 2$ (for this illustrating example using short time series, and rounding up values to a one-digit mantissa, it happened that $k_1$ and $k_2$ are equal. Real time series are usually much longer).

### 3.4 Complexity Considerations

The TSAX representations for all the time series in the dataset are to be computed and stored offline. The length of a TSAX representation is $2m$, whereas that of SAX is $m$. However, the actual additional storage requirement is much less than that because the additional storage requirement of TSAX, which is the trend part, uses Boolean variables, which require one bit each, whereas the symbols used to represent SAX (and the symbolic part of TSAX), use characters, which require 8 bits each. What is more, in many cases the trend of several successive segments of a time series is the same. This gives the possibility of using the common lossless compression technique *Run Length Encoding* (RLE) [5].

As for the computational complexity when applying equation (3). This is also a tiny additional cost compared with SAX, because as we can see from equation (3), the two additional computations are $rew \times k_1$ and $pen \times k_2$. Each of them requires one multiplication operation, and $m$ logical comparison operations, which have small computational latency, especially compared with the other very costly operation (sqrt) required to compute equation (1) in SAX.

So as we can see, the additional storage and computational costs that TSAX adds are quite small. During experiments, we did not notice any difference in computational time.

## 4  Experiments

We tested TSAX against SAX in a TSC task using 1NN, which we presented in Section 1. We avoided any choice that could bias our method, so we chose the time series archive which is suggested by the authors of SAX to test time series mining methods: *UCR Time Series Classification Archive* [3]. This archive contains 128 time series datasets. To avoid "cherry picking", the archive managers suggest testing any new method on all the datasets, which is not possible to present here for space limitations. However, we decided to do the second best thing to avoid favoring our method; because the datasets are arranged in alphabetical order of their names, we simply decided to test both TSAX and SAX on the first 50 datasets of the archive. Because the name of a dataset is by no means related to its nature, this choice is random enough to show that we did not choose to experiment on datasets that might bias TSAX over SAX. Besides, in other experiments, which we do not report here, on other randomly chosen datasets from the archive, we got similar results.

The datasets in the archive are divided into test and train, but neither SAX nor TSAX need training (we will show later why TSAX does not need training), so the two methods were applied directly to the test datasets, which is the same protocol used to validate SAX in the original paper.

As for the value of the alphabet size $\alpha$ for both SAX and TSAX, SAX can be applied to any value between 3-20, but in [13] Lin et al, the authors of SAX say concerning the choice of the alphabet size "a value of 3 or 4 works well for most time series datasets. In most experiments, we choose α = 4 for simplicity", so we too chose α = 4 (We conducted other experiments, not shown here, for α = 3, and we got similar results). We also chose the same compression ratio (i.e. $n/m$ in equations (1) and (3)).

**Table 2.** The datasets, the number of time series in each dataset, the number of classes, the length of the time series, the classification errors of SAX, and the classification errors of TSAX. The best result for each dataset is shown in boldface printing.

| Dataset | Nr. of TS | Nr. of Classes | Length | SAX | TSAX |
|---|---|---|---|---|---|
| Adiac | 391 | 37 | 176 | 0.974 | **0.852** |
| ArrowHead | 175 | 3 | 251 | 0.571 | **0.309** |
| Beef | 30 | 5 | 470 | 0.667 | **0.167** |
| BeetleFly | 20 | 2 | 512 | **0.250** | 0.300 |
| BirdChicken | 20 | 2 | 512 | **0.350** | 0.500 |
| Car | 60 | 4 | 577 | 0.550 | **0.350** |
| CBF | 900 | 3 | 128 | **0.236** | 0.579 |
| ChlorineConcentration | 3840 | 3 | 166 | 0.742 | **0.495** |
| CinCECGTorso | 1380 | 4 | 1639 | 0.304 | **0.223** |
| Coffee | 28 | 2 | 286 | 0.464 | **0.250** |
| Computers | 250 | 2 | 720 | 0.564 | **0.444** |
| CricketX | 390 | 12 | 300 | **0.513** | 0.759 |
| CricketY | 390 | 12 | 300 | **0.531** | 0.741 |
| CricketZ | 390 | 12 | 300 | **0.485** | 0.769 |
| DiatomSizeReduction | 306 | 4 | 345 | 0.696 | **0.160** |
| DistalPhalanxOutlineAgeGroup | 139 | 3 | 80 | 0.717 | **0.240** |
| DistalPhalanxOutlineCorrect | 276 | 2 | 80 | 0.412 | **0.317** |
| DistalPhalanxTW | 139 | 6 | 80 | 0.722 | **0.320** |
| Earthquakes | 139 | 2 | 512 | 0.317 | **0.180** |
| ECG200 | 100 | 2 | 96 | 0.220 | **0.190** |
| ECG5000 | 4500 | 5 | 140 | 0.195 | **0.077** |
| ECGFiveDays | 861 | 2 | 136 | 0.475 | **0.236** |
| ElectricDevices | 7711 | 7 | 96 | 0.879 | **0.584** |
| FaceAll | 1690 | 14 | 131 | 0.571 | **0.463** |
| FaceFour | 88 | 4 | 350 | **0.205** | 0.330 |
| FacesUCR | 2050 | 14 | 131 | 0.476 | **0.447** |
| FiftyWords | 455 | 50 | 270 | **0.415** | 0.499 |
| Fish | 175 | 7 | 463 | 0.851 | **0.280** |
| FordA | 1320 | 2 | 500 | 0.377 | **0.332** |
| FordB | 810 | 2 | 500 | 0.451 | **0.444** |
| GunPoint | 150 | 2 | 150 | 0.260 | **0.167** |
| Ham | 105 | 2 | 431 | 0.486 | **0.419** |
| HandOutlines | 370 | 2 | 2709 | 0.283 | **0.187** |
| Haptics | 308 | 5 | 1092 | 0.718 | **0.640** |
| Herring | 64 | 2 | 512 | **0.406** | 0.469 |
| InlineSkate | 550 | 7 | 1882 | 0.800 | **0.769** |
| InsectWingbeatSound | 1980 | 11 | 256 | 0.494 | **0.480** |
| ItalyPowerDemand | 1029 | 2 | 24 | 0.459 | **0.070** |
| LargeKitchenAppliances | 375 | 3 | 720 | 0.643 | **0.592** |
| Lightning2 | 61 | 2 | 637 | **0.311** | 0.492 |
| Lightning7 | 73 | 7 | 319 | **0.507** | 0.808 |
| Mallat | 2345 | 8 | 1024 | 0.668 | **0.309** |
| Meat | 60 | 3 | 448 | 0.667 | **0.533** |
| MedicalImages | 760 | 10 | 99 | 0.599 | **0.539** |
| MiddlePhalanxOutlineAgeGroup | 154 | 3 | 80 | 0.730 | **0.292** |
| MiddlePhalanxOutlineCorrect | 291 | 2 | 80 | 0.647 | **0.430** |
| MiddlePhalanxTW | 154 | 6 | 80 | 0.601 | **0.431** |
| MoteStrain | 1252 | 2 | 84 | 0.277 | **0.165** |
| NonInvasiveFetalECGThorax1 | 1965 | 42 | 750 | 0.908 | **0.789** |
| NonInvasiveFetalECGThorax2 | 1965 | 42 | 750 | 0.871 | **0.704** |
| | | | | 11/50 | **39**/50 |

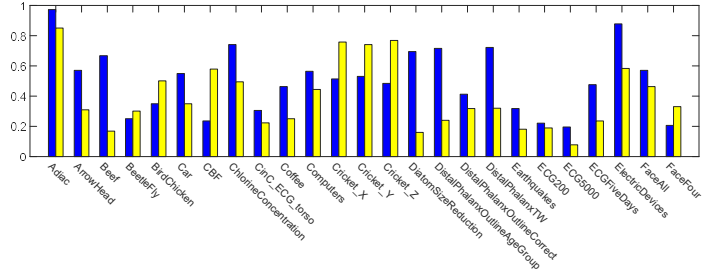

(a)

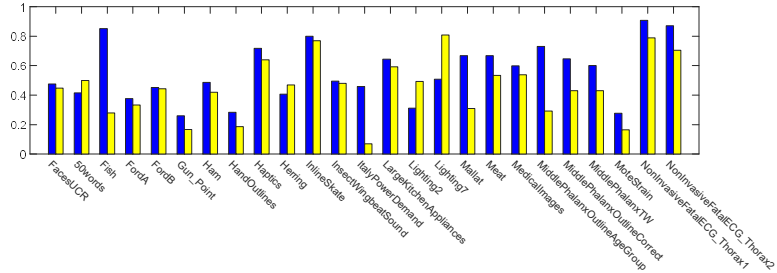

(b)

**Fig. 3.** Bar chart comparisons of the classification errors between TSAX (in yellow) and SAX (in blue) on the first 25 datasets (a) and the second 25 datasets. The figure shows the superior performance of TSAX over SAX.

TSAX has two parameters that SAX does not have, these are $rew$ and $pen$. In order to obtain the best performance of TSAX, these two parameters need to be trained on each dataset. However, because our intention was to propose a method that has the same simplicity as SAX, we chose the same value for $rew$ and $pen$ for all the datasets. And although even in this case, we could at least test different values on different datasets to decide which ones work best, we decided to push our method to the limit, so we chose, rather arbitrarily, $rew = -1$ and $pen = 1$. The rationale behind this choice is that we thought when two segments have the same trend this is equivalent to the distance between two symbols (In Table 1: 1.34 - 0.67, rounded to the closest integer, and taken as a negative value), so we chose $rew = -1$ and we simply chose $pen$ to be the opposite of $rew$. We know this choice is rather arbitrary, which is intentional, because we wanted to show that the performance of TSAX is independent of the choice of the parameters $rew$ and $pen$, so as such, they are treated as constants.

Table 2 and Fig. 3 show the classification errors of the 50 datasets on which we conducted our experiments for TSAX and SAX. As we can see, TSAX clearly outperforms SAX in TSC as it gave better results, i.e. a lower classification error, in 39 out of the 50 datasets tested, whereas SAX gave a lower classification error in 11 datasets only. In some datasets, like (Beef), (DiatomSizeReduction), (DistalPhalanxOutlineAgeGroup), (ItalyPowerDemand), the gain in applying TSAX is substantial, whereas the only dataset we see the substantial gain in applying SAX is (CBF) (which is a synthetic dataset).

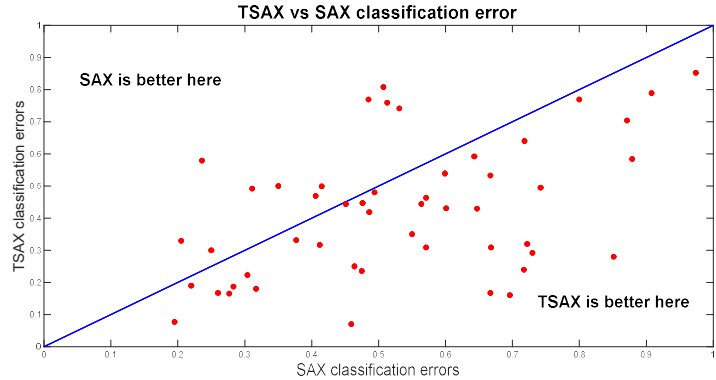

**Fig. 4.** Comparison of the classification errors of TSAX versus SAX. The bottom-right region is where TSAX performs better, and the top-left region is where SAX performs better.

An interesting phenomenon that we notice is that for certain "groups" of similar datasets, such as (CricketX), (CricketY), (CricketZ), and (DiatomSizeReduction), (DistalPhalanxOutlineAgeGroup), (DistalPhalanxOutlineCorrect), (DistalPhalanx-TW), or (MiddlePhalanxOutlineAgeGroup), (MiddlePhalanxOutlineCorrect), (MiddlePhalanxTW), the performance is usually the same. We believe this validates our motivation that for some time series datasets, trend information is quite crucial in performing TSC, whereas for fewer others it is not important.

Finally, in Fig. 4 we show, on one plot, a comparison of the classification errors for TSAX versus SAX on all 50 datasets. This global view of the outcomes of the experiments summarizes our above mentioned findings. The bottom-right region of the figure shows where the classification errors of TSAX are lower than those of SAX, whereas the top-left region of the figure shows where the classification errors of SAX are lower than those of TSAX

## 5   Conclusion

In this paper we presented TSAX, which is a modification of SAX - one of the most popular time series representation methods. TSAX enhances SAX by providing it with features that capture segment trending information. TSAX has almost the same simplicity and efficiency of SAX, but it has a better performance in time series classification, which we showed through extensive experiments on 50 datasets, where

TSAX gave a lower classification error than SAX on 39 of these datasets. For some datasets the improvement in performance was substantial.

In the experiments we conducted we meant to focus on simplicity. But we can even improve the performance of TSAX further by finding the best values for the parameters *rew* and *pen*.

Although we tested TSAX in a time series classification task only, we believe, which is one direction of future work, that TSAX can give better results than SAX in other time series mining tasks as well.

Like SAX, TSAX compares two time series segment by segment, we are investigating comparing two time series on a several-segment basis at a time. This will be of interest on both the symbolic and the trend parts of the TSAX representation. We are particularly interested in integrating some techniques from string comparisons that are used in the bioinformatics community.